# Iterative Thresholded Bi-Histogram Equalization for Medical Image Enhancement

Qadar Muhammad Ali
Key Laboratory of Intelligent Information Processing, Institute of Computing Technology, Chinese Academy of Sciences, Beijing 100190, China

Zhaowen Yan
School of Electronic Information Engineering Beihang University, Beijing 100191, China

Hua Li
Key Laboratory of Intelligent Information Processing, Institute of Computing Technology, Chinese Academy of Sciences, Beijing 100190, China

## ABSTRACT
Enhancement of human vision to get an insight to information content is of vital importance. The traditional histogram equalization methods have been suffering from amplified contrast with the addition of artifacts and a surprising unnatural visibility of the processed images. In order to overcome these drawbacks, this paper proposes interative, mean, and multi-threshold selection criterion with plateau limits, which consist of histogram segmentation, clipping and transformation modules. The histogram partition consists of multiple thresholding processes that divide the histogram into two parts, whereas the clipping process nicely enhances the contrast by having a check on the rate of enhancement that could be tuned. Histogram equalization to each segmented sub-histogram provides the output image with preserved brightness and enhanced contrast. Results of the present study showed that the proposed method efficiently handles the noise amplification. Further, it also preserves the brightness by retaining natural look of targeted image.

## 1 . INTRODUCTION
Histogram equalization is a common and foremost effective technique used for image contrast enhancement. It simply uses cumulative density function (CDF) to redistribute the gray levels of input image. One of the drawbacks of this technique is that it does not consider the input image's brightness preservation that is the most important factor for processing of consumer electronics, medical, SAR images. Brightness preservation was first introduced by Kim et.al, 1997 [1], followed by Wang et.al, 1999 [2] and Chen et.al, (2003) [3]. Ooi et.al, (2009) [4] introduced clipping process in brightness preserving bi-histogram equalization (BBHE). Lim et.al (2013) [5] improved the BBHE method by introducing six threshold plateau limits for segmented histogram. Zuo et.al. (2012) [6] introduced an idea of histogram partition by minimizing the intera-class variance. Kuldeep et.al (2013) [7] divided the histogram by using exposure threshold values and then applied the clipping process followed by equalizing each partition.

This paper presents three different ways to segment the histogram, which comprises of searching the best suited threshold, using the multi-thresholding concept [8], and mean value based partition respectively. Plateau limits for both sub-histograms are calculated based on the probability density function of input histogram. Later clipping process clips the sub-histogram based on their cumulative redistribution of histogram. After successful partitions, histogram equalization is applied to each partition independently. This algorithm is applicable to grayscale images including natural, medical, satellite image data etc. The proper enhancement and its time complexity make it suitable for real time applications.

The paper is organized into four sections. Section II explains briefly the previous work done in image enhancement with respect to brightness preservation and contrast enhancement. Section III introduces the proposed method with detailed algorithmic steps. Section IV is divided into subjective and objective analysis of the proposed methods with detailed discussion on algorithm results. Acknowledgements and conclusive remarks are presented in section V and VI respectively.

## 2 . PREVIOUS WORK
Some basic definitions, here $I$ would represents input image, $Y$ would represents output image, $[X_l \ X_u]$ represents the lower and upper limit of boundaries for image $X$.

Preservation of brightness and contrast are considered as major reasons for poor quality in enhancement of images [9]. Histogram equalization can introduce a significant change in brightness of an image, which hinders the direct application of the histogram equalization. More fundamental reasons behind the limitations of the histogram equalization are that the histogram equalization does not take the mean brightness of an image into account [1]. Kim et.al, (1997) [1] proposed brightness preserving bi-histogram equalization (BBHE) that aims to address the draw backs of HE as explained earlier, it divides the histogram of image based on its mean value and then equalizes each sub-histogram independently [7].

Mean brightness of the image equalized by the BBHE locates in the middle of the input mean and the middle gray level. Note that the output mean of the BBHE is a function of the input mean brightness. This fact clearly indicates that the BBHE preserves the brightness compared to the case of a typical histogram equalization where the output mean is always the middle gray level [1].

Wang et.al, (1999) [2] proposed that if histogram could be divided based on gray level distribution value, it can preserve more brightness. Entropy of image could be defined as that is used to represent the richness of details in the image. The segmentation entropy will achieve the maximum value when the two sub-images have equal area, it is sure that the average luminance of the original image could be kept from significant shift especially for the large area of the image with the same gray level [2]. Chen et.al, (2003)[3] proposed a novel extension of brightness preserving bi-histogram equalization (BBHE). This algorithm states that separating the histogram based on a threshold value could yield a minimum mean brightness error. There are some cases that could not be handled by HE, BBHE, DSIHE.

Procedural steps of Minimum mean brightness error bi-histogram equ-alization (MMBEBHE) are given below:

1. Calculate the AMBE for each of the threshold level.

2. Find the threshold level, $I_T$ that yield minimum MBE,

3. Separate the input histogram into two based on the $I_T$ found in step 2 and equalize them independently as in BBHE.

From the results of the proposed algorithm, it's clearly shown that the brightness preservation (mean brightness) has increased and yielded a more natural enhancement. Ooi et.al, (2009) [4] proposed an algorithm that successfully preserves the brightness maintaining the automatic selection of parameter. However, there are some methods that require the manual selection of parameters, Bin underflow and bin overflow histogram equalization (BUBOHE) [10], Wieghted threshold histogram equalization (WTHE) [11] and Gain controllable clipped histogram equalization (GC-CHE) [12] etc. The algorithm first applies the BBHE process as the first step of segmenting the histogram, then it selects the plateau limits in both parts of the histogram, finally it simply clips and equalizes each histogram independently. The beauty of this algorithm is the processing time and good natural enhancement. The image $I$ is segmented based on mean value $I_m$ and divided into two parts $I_L$ and $I_U$ same as in BBHE, then plateau thresholds are calculated.

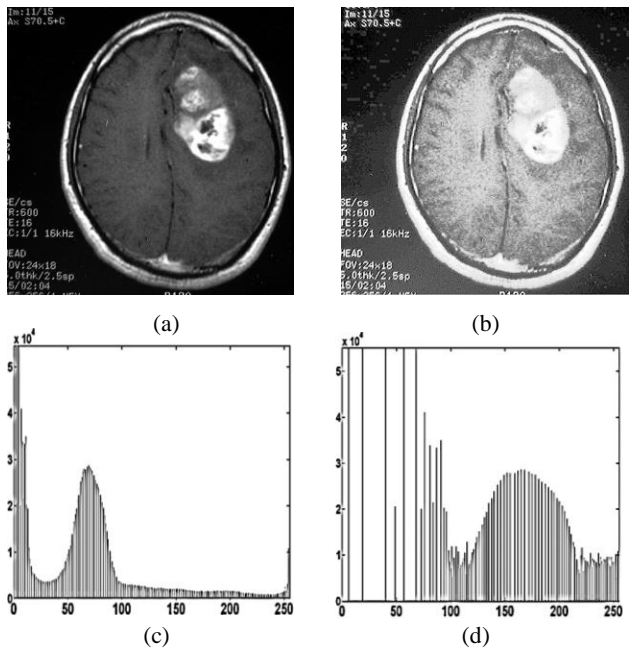

Fig 1: Effect of HE method for brain image  a) Original image  b) Processed image  c) Histogram of Original image  d) Histogram of Processed image

Plateau limits are purpose on to limit the intensity saturation of the output image. Finally, HE is applied to the output image. Bi-histogram equalization with plateau limit (BHEPL) does not redistribute the clipped portions back into the modified histogram. As a consequence, BHEPL is simple to be implemented and requires less hardware.

Lim et.al (2013) [5] proposed a simple and effective idea based on BBHE and BHEPL. The algorithm simply performs BBHE segmentation to divide the histogram into two parts, then the plateau limits are calculated from respective sub-histograms, and which are used to modify those sub-histograms. Histogram equalization is then separately performed on the two sub histograms to yield a clean and enhanced image. The algorithm is composed of mainly three parts, histogram segmentation, modification and transformation. After the dividing the histogram same as in BBHE the methods follow the histogram modification by first selecting the plateau limits. Finally, HEd image could be obtained. The performance of algorithm in contrast enhancement and addressing low gray level images makes it useful for the medical image enhancement and applicable to consumer electronics, SAR, video cameras, etc.

Zuo et. al.(2013) [6], proposes an algorithm which segments the histogram into two parts based on Otsu (1979) [8] thresholding that limit the range of equalized image. The algorithm assumes that the image to be thresholded contains two classes of pixels (e.g., foreground and background) then calculates the optimum threshold separating those two classes so that their intra-class variance is minimal. It exhaustively searches for the threshold that minimizes the intra-class variance, defined as a weighted sum of variances of the two classes

$$\sigma^2(X_T) = W_L(E(X_L) - E(X))^2 + W_U(E(X_U) - E(X))^2 \qquad (1)$$

$E(X_L)$ and $E(X_U)$ are the average brightness of two sub-images thresholded by $X_T$. $E(X)$ is the mean brightness of the input image. $W_L$ and $W_U$ are the weights that exhibit two classes of pixels. Later HE process is applied to equalize each partition independently. The algorithm has its beauty in the preserving the brightness and enhancement of the contrast. Its simplicity makes it applicable to real time applications.

## 3. THE PROPOSED METHOD
Bi-histogram equalization based methods could prominently enhance the image with good brightness preservation to some extent, but the images obtained by these methods look unnatural.

Due insufficient parameters that process histogram a proper and natural enhanced image could not be obtained. This work is done in Bi- histogram equalization domain to evolve with a better way to preserve brightness and enhance contrast. Hence, which will be applicable to real time applications in its processing of time and simplicity.

Firstly it has been focused on the point that how to decompose the image, the decomposition process is involved the threshold selection criteria, as image segmentation process is first step towards the goal in preserving the brightness. Threshold selection is firstly a searching criteria and automated process that chooses the best suitable threshold. The other way to select the threshold is using [8], two threshold are selected using Otsu's method that suits to target image. To control the rate of enhancement, an automated plateau limit selection and histogram clipping processing are applied. Finally processed image is equalized by using histogram equalization. Method includes following modules, histogram segmentation, histogram clipping, and transformation.

### 3.1 Histogram Segmentation
Many methods exist in literature that is aiming at partitioning the histogram. All of these aimed to preserve the brightness and optimize the entropy of the output image. However entropy optimization is a critical task to be obtained by only partitioning the histogram [13]. To preserve the natural look with better statistical parameter values, the histogram is clustered into classes. Each sub-image has relative class correspondence that minimizes the brightness shift because of histogram equalization. Different ways of histogram segmentation have been implemented first of these methods are searching the optimal threshold. The detailed steps of the proposed methods are as follows:

a). Find the optimal threshold $\hat{X}$ through searching

The aim of the optimal threshold is to minimize the mean brightness error that could be defined as

$$\hat{X} = arg \min_{x}(E(Y) - E(I)) \quad (2)$$
$$x = x_i \ \forall \ i = 1,2,3 \dots, N \quad (3)$$

Where $N$ is the maximum size of image matrix e.g. of 256*256 image value of $N = 256$. Where $x$ is the selected threshold value from image matrix of 256 size.

As the finding thresholding is an iterative search procedure that aims at reducing the mean error. The segmentation process is illustrated in the Fig.3. Flow graph of the proposed method is presented as follows in Fig.2. The graph in Fig. 3 resulted from the process that the histogram of the input image is segmented based on mean value. $X_T$ is considered as mean value of the image. Resultant sub-histograms are processed further using clipping process as described in Fig.2.

b). The second kind of threshold of selection criteria that is adopted here is to select threshold values based on multi-threshold [8], two thresholds have been selected that divides that histogram into three parts. The purpose here is to determine the threshold that minimizes the weighted within-class variance. It's defined as follows

$$\sigma^2(X_t) = p_l(I_k)\big(E(I_L) - E(X)\big) + p_u(I_k)\big(E(I_U) - E(X)\big) \quad (3)$$

So the threshold calculated could be written as

$$X_T = \underset{X_t}{argmax}\{\sigma^2(X_t), t = 0,1,\dots,L-1\} \quad (4)$$

For three thresholding, there are two thresholds are assumed $0 \leq X_{T1} < X_{T2} < L-1$ with three separated classes $M_0 = \{1 \dots X_{T1}\}$, $M_1 = \{X_{T1}+1 \dots X_{T2}\}$, and $M_2 = \{X_{T2}+1 \dots L-1\}$. Hence, the optimal set of thresholds is selected by maximizing the $\sigma^2$

$$\sigma^2(X_{TA}, X_{TB}) = \underset{0 \leq X_{T1} < X_{T2} < L-1}{argmax} \sigma^2(X_{T1}, X_{T2}) \quad (5)$$

$X_{TA}, X_{TB}$ are the two optimal threshold which are then used to segment the histogram.

c). The third kind of threshold selection is using mean value with the addition of clipping module, mean value could be calculated as follows:

$$X_T = \sum_{k=0}^{L-1} I_k \times n_k / N \quad (6)$$

Where $I_k$ is the kth gray-level, $n_k$ are number of pixels of gray-level k, $N$ total number of pixels in input test image. $X_T$ is similarly the threshold value to segment the histogram into two parts. The threshold that is obtained from above equation is used to segment the histogram as follows

Let's denote $I_x$ as the input threshold value of image $I$, where $I_x \in \{I_0, I_1, \dots I_{L-1}\}$. Image is decomposed into two parts using $I_x$ into $I_L$ & $I_U$, $I$ could be written as

$$I = I_L \cup I_U \quad (7)$$

Where

$$I_L = \{I(X_L, X_U) | I(X_L, X_U) < I_x, I(X_L, X_U) \in I\} \quad (8)$$
$$I_U = \{I(X_L, X_U) | I(X_L, X_U) \geq I_x, I(X_L, X_U) \in I\} \quad (9)$$

Where $[X_L, X_U]$ have same meaning as $[0, L-1]$ Note that the sub-image $I_L$ is in the range $\{I_0, I_1, \dots I_x\}$ and $I_U$ is in range of $\{I_{x+1}, I_{x+2}, \dots I_{L-1}\}$. Now similarly like BBHE define the respective probability density functions for both sub-images $I_L$ & $I_U$ are denoted as $p_l$ and $p_u$.

$$p_l(I_k) = \frac{n_L^k}{n_L} \quad (10)$$

Where $k = \{0,1,2 \dots x\}$

$$p_u(I_k) = \frac{n_U^k}{n_U} \quad (11)$$

Where $k = \{x+1, x+2 \dots L-1\}$, in which $n_L^k$ and $n_U^k$ represent the respective number of $I_k$ in $I_L$ & $I_U$ and $n_L, n_U$ represent the total number of samples in $I_L$ & $I_U$. It could be noted that $n_L = \sum_{k=0}^{x} n_L^k$ and $n_U = \sum_{k=x+1}^{L-1} n_U^k$.

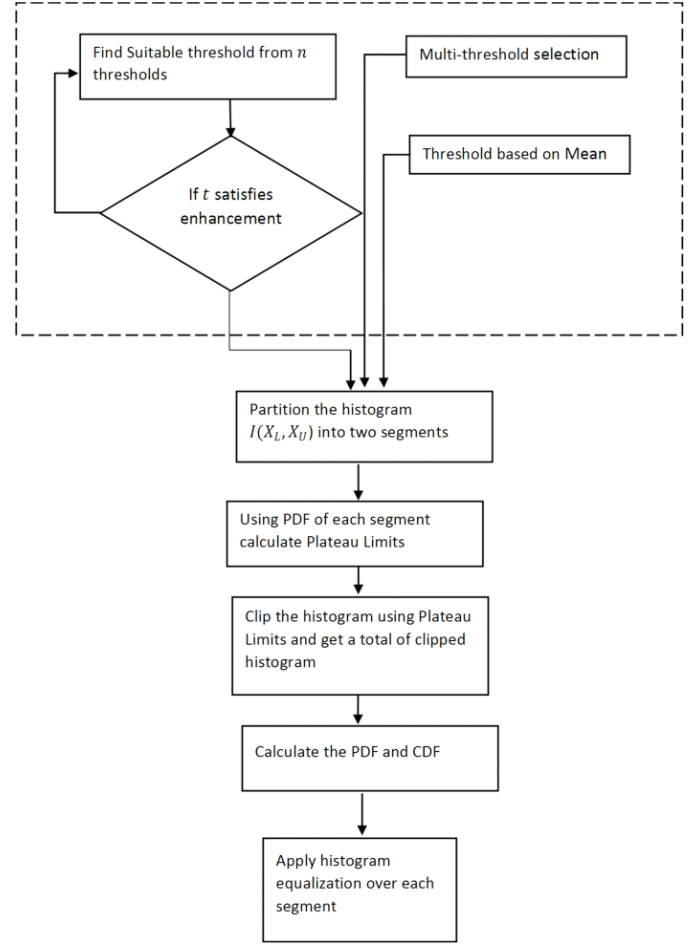

Fig 2: Flow graph of proposed method

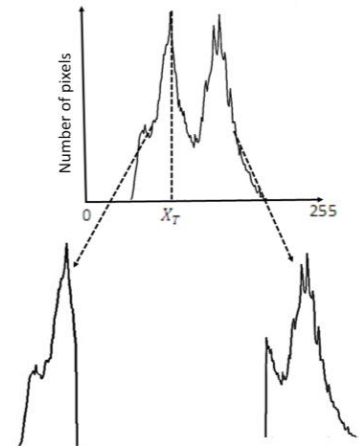

Fig 3: Histogram Segmentation process

## 3.2 Histogram Clipping
A probability density function is used in selecting the thresholds as plateau limits for the segmented histogram using optimal threshold value. The histogram of the $I_L$ & $I_U$ could be denoted as $H_L$ and $H_U$, Hence left and right plateau limits are could determine as follows

$$T_L = \frac{1}{I_x + 1}\sum_{k=0}^{I_x} p_l(k) \qquad (12)$$

$$T_U = \frac{1}{(L-1) - I_x}\sum_{k=I_x+1}^{L-1} p_u(k) \qquad (13)$$

$T_L$ is actually an average of $p_l$, after that to control the rate of enhancement these limits are used to clip the histogram from specified threshold obtained from $T_L$ and $T_U$.

The clipped histograms are denoted as $H_{LL}$ and $H_{UL}$, the process of clipping which is shown in Fig.4 could be performed by following equations given below

$$H_{LL} = \begin{cases} H_L(i) & if\ H_L(i) \leq T_L \\ T_L & elsewhere \end{cases} \qquad (14)$$

$$H_{UL} = \begin{cases} H_U(i) & if\ H_U(i) \leq T_U \\ T_U & elsewhere \end{cases} \qquad (15)$$

The total of the clipped histogram could be calculated as

$$W1 = \sum_{k=0}^{L-1} H_{LL}(k) \qquad (16)$$

And

$$W2 = \sum_{k=I_x+1}^{L-1} H_{UL}(k) \qquad (17)$$

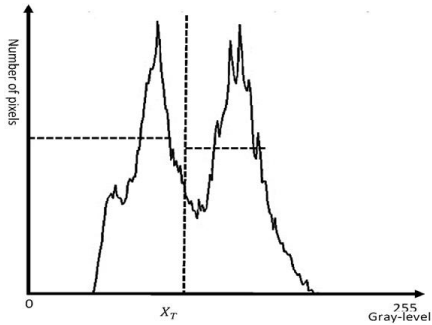

Fig 4: Clipping Process

After that probability density of processed histogram is used to calculate cumulative density function that follows as

$$C_L = \frac{1}{W1}\sum_{k=0}^{L-1} p_{ll}(I_k) \qquad (18)$$

$$C_U = \frac{1}{W2}\sum_{k=x+1}^{L-1} p_{ul}(I_k) \qquad (19)$$

$p_{ll}$ and $p_{ul}$ could be calculated same way as stated above, the cumulative density should be $C_L(I_x) = 1$ and $C_L(I_{L-1}) = 1$

### 3.3 Transformation

As this method is based on histogram equalization, cumulative density function is used to transform to allocate the new range of intensity values. The transform function to equalize the processed image as an output image is given as follows

$$f_L(i) = I_0 + (I_x - I_0)C_L(i) \qquad (20)$$

$$f_U(i) = I_{x+1} + (I_{L-1} - I_{x+1})C_U(i) \qquad (21)$$

Finally the output is

$$Y = Y(X_l, X_u) = f_L(Y_L) \cup f_U(Y_U) \qquad (22)$$

$$Y(X_l, X_u) = \begin{cases} I_0 + (I_x - I_0)C_L(i) & if\ i \leq I_x \\ I_{x+1} + (I_{L-1} - I_{x+1})C_U(i) & else \end{cases} \qquad (23)$$

$Y(X_l, X_u)$ is the final histogram equalized output image. Following section explains results of experiment on qualitative and quantitative basis.

## 4. RESULTS AND DISCUSSION

Target images are medical images of different parts of body. All results are compared with existing techniques [14] [1] [2] [7] [20] [5][6] for assessing their relative performance. The target images used in their experiments are gray-scale medical images. To validate the performance of algorithm different kind of medical images are selected. All images are 256 ×256 in 2 dimensions, there are total 120 images have been tested experimentally using the proposed and existing algorithms, all algorithms have been tested on corei7 desktop pc. The proposed algorithm has been divided into three parts based on the segmentation of histogram, hence three different results have been obtain. Based on the results and discussion from performance overview best algorithm is also analyzed. All dataset is self-made, collected from hospitals, and internet resources etc.

### 4.1 Qualitative (Subjective) Analysis

In order to assess the quality and appropriateness of the above shown images, AMBE [3] have been computed. It has stated in the literature that mean brightness error is of vital importance for the quality assessment of enhanced images. The standard for the mean brightness error is that as low as brightness error, the more is the preservation of brightness with a quality enhanced image. The values obtained for each image is given in the tables above, the results are also given to the other methods, HE, BBHE, DSIHE, MMBEBHE, Brightness preserving histogram equalization with plateau limits (BHEPL),Range Limited Bi-Histogram Equalization (RLBHE) etc. From the set of 120 images four of the medical images are presented for comparison purposes. Analyzing the Table-5 reveals that the absolute difference between processed and original image is less than other methods in competition. The more close look to the visual quality of Fig.5, 6, 7, 8 image, there is over-enhancement effect in the methods HE, BBHE, DSIHE, and also have generated the noise amplification effect, whereas for the MMBEBHE, there is no over-enhancement, and very good brightness preserved with value 1.05, but in fact the algorithm suffers effect of noise and no better visual quality than proposed algorithm. In Fig.5, 6, 7, 8 (i) BHEPL also performs better than BBHE and DSIHE but its brightness preservation is no more than MMBEBHE in this case of the medical image. However algorithm suffers a little over-enhancement, whereas new histogram equalization for brightness preserving and contrast enhancement (NHEBP) performs well in enhancement, but there is not much effect on the quality because of the clipping process that's controlled enhancement beyond the need of enhancement as seen in the Fig.5, 6, 7, 8 (j). Whereas in Fig.5, 6, 7, 8 (k) RLBHE have introduced the washout appearance with intensity saturation in the image. Hence, as it clear from the result of AMBE, proposed method is performing better with fine visual quality and minimized AMBE value as compared to other

23

algorithms. Similarly parameter like standard deviation with maximum attainable value presents the better image quality [15] [14] which is shown in the Table-6. Entropy measures the richness of information, higher its value higher the information content [13] which is shown in the Table-7. Where Table-8 presents the peak signal to noise ratio (PSNR) value that should be between 20db-30db as stated in [21]. The results for the four images tells that images are enhanced with quite better quality as seen from the PSNR value. For better look of image the value Universal image quality index (UIQI) should be closed to unity [17] which is shown in Table-9. Where Table-10 presents the enhancement error (EME) that states that it should be minimum for original and processed images. Structural similarity index (SSI) is the parameter employed to asses quality and natural look of processed image, closer to unity is the recommended value of SSI [19] [22]. For the head MRI image Fig.6 (b-d) the proposed algorithm has shown a fine visual quality showing with rich information content that is also proved from all of the statistic parameters AMBE, SD, PSNR etc. Closer look to the jaw in this head MRI Fig.6 (b-d) shows fine visual details that other methods without generating artifacts and noise effect. Whereas Fig.7 low quality image, ITSBPL and MSBPL in Fig.7 (b-c) have shown better brightness preserved with a good quality image and MVSBPL in Fig.7 (d) has made the image visually more bright within an acceptable range of AMBE. In Fig.8 image of the kidney is a complex structure with small veins visible inside. The goal is to enhance the image in such a way to clarify the richness of details. Hence that is achieved by employing MVSBPL in Fig.8 (d) as shown. Whereas existing algorithm does not enhance image properly and have created some noise patches and over-enhancement taking into account HE, BBHE, DSIHE, in Fig.8 (e-g). The algorithms like MMBEBHE, BHEPL, NHEBP does not effect original contrast or brightness for Fig.8 specifically. In Fig. 8 (k) RLBHE performs better as compared to HE, BBHE, DSIHE, MMBEBHE, BHEPL, and NHEPB by fine visualization of overall image quality, but still over-enhancement affect is inevitable. Ranking of enhancement results, MVSBPL as first among other two approaches, whereas ITSBPL and MSBPL are ranked as second and third respectively.

## 4.2 Quantitative (Objective) Analysis
There are different parameter that have been selected the parametric measurement on the quality and performance measurement and comparison with already defined techniques

*4.2.1 Absolute mean brightness error (AMBE)*
This difference of mean brightness between input and output image. This parameter helps to figure out the quality of image in brightness preservation that is major and first foremost parameter in image quality assessment [3]. It could be defined as

$$AMBE(I,Y) = |E(I) - E(Y)| \quad (24)$$

Where $E(I)$ is the input image's brightness and $E(Y)$ is the output image's mean brightness. Lower value means good brightness preservation.

*4.2.2 Standard deviation (SD)*
Standard deviation is basic parameter is used in image quality measurment, it could be denoted as $\sigma$ and could be defined by following equation:

$$\sigma = \sqrt{\sum_{k=0}^{L-1}(Y_k - I_m)^2 \times p_k(Y_k)} \quad (25)$$

Where $Y_k$ is the resultant image and $I_m$ is the mean brightness of the equalized image, $p_k(Y_k)$ is the probability density of $Y_k$. Higher the value of SD, better are the enhancement results. Higher standard deviation sometime does not mean always that contrast is enhanced with better quality[15] [14].

*4.2.3 Entropy*
Measure the richness of information in the image [14]. Higher the value of entropy, higher the detailed information image contains, it could be defined as follows

$$Ent(Y_k) = -\sum_{k=0}^{L-1} p_k(Y_k).log_2\, p_k(Y_k) \quad (26)$$

Where $p_k(Y_k)$ is the PDF of the output image, and $Ent(Y_k)$ exhibits the entropy of resultant image [15].

*4.2.4 Peak signal to Noise ratio (PSNR)*
In order to assess the pixels distribution and their appropriateness in the output image, PSNR is the best suited parameter as defined in [16]

$$PSNR = 10log_{10}[\frac{(L-1)^2}{MSE}] \quad (27)$$

MSE is called as the root mean square error that could be defined as

$$MSE = \sum_{X_l}\sum_{X_u}\frac{|I(X_l,X_u) - Y(X_l,X_u)|^2}{N} \quad (28)$$

Where $I(X_l,X_u), Y(X_l,X_u)$ are the corresponding pixel values in respective input and output images and $N$ are the total pixel values.

*4.2.5 Universal image quality index (UIQI)*
This is used in the process of evaluating the natural appearance of the contrast enhanced image. This method is used to assess the quality taking into account of natural look for different histogram equalization based methods. UIQI could be defined as follows

$$UIQI = \frac{4\sigma_{ab} \times I_m \times Y_m}{\sigma_a^2 + \sigma_b^2[(I_m)^2 \times (Y_m)^2]} \quad (29)$$

Where $I_m$ and $Y_m$ are the mean intensity level for the both input and output images. $\sigma_{ab}^2, \sigma_a^2, \sigma_b^2$ are defined as follows

$$\sigma_a^2 = \frac{1}{N-1}\sum_{k=1}^{N}(I_k - I_m)^2,\ \ \sigma_b^2 = \frac{1}{N-1}\sum_{k=1}^{N}(Y_k - Y_m)^2,\ \sigma_{ab}^2 = \frac{1}{N-1}\sum_{k=1}^{N}(I_k - I_m)(Y_k - Y_m) \quad (30)$$

There are three different kinds of relation have formed from above equations that could be called as loss of correlation, luminance distortion, and contrast distortion. For better preservation of natural appearance the value of the UIQI should be closer to unity [17].

*4.2.6 Enhancement Error (EME)*
It's the parameter used for the quantitative measurement of for an image $I_k$ of size $\times N$, it is defined by following equation

$$EME(I_k) = EME_\Phi(I_k) = \frac{1}{k^2}\sum_{n=1}^{k}\sum_{m=1}^{k}\frac{(\max(I_k([n,m]))}{(\min(I_k([n,m]))} \quad (31)$$

Where $n,m$ signifies the chunk of the image $I_k$, and the image is divided by $k^2$ blocks with $L \times L$ as assigned size and $k = [N/L], [.]$

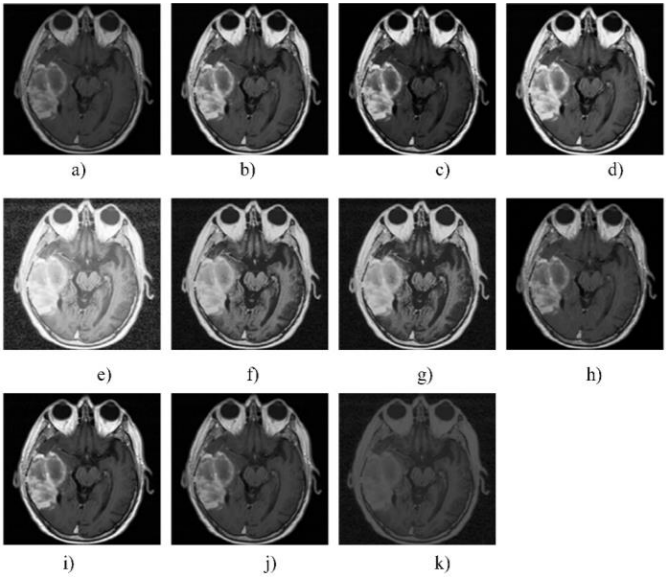

Figure 5: Brain MRI image enhancement   a) original Image b) ITSBPL c) MSBPL d) MVSBPL e) HE f) BBHE   g)  DSIHE  h) MMBEBHE   i) BHEPL j) NHEBP k) RLBHE

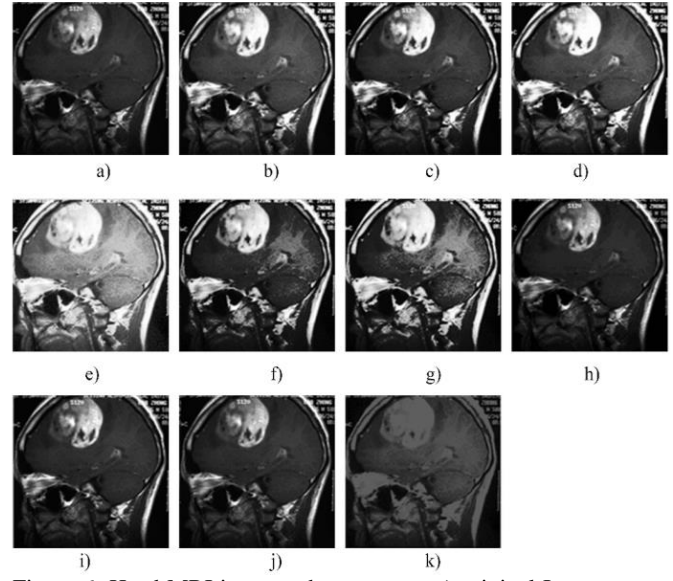

Figure 6: Head MRI image enhancement   a) original Image b) ITSBPL c) MSBPL d) MVSBPL e) HE f) BBHE g) DSIHE   h) MMBEBHE   i) BHEPL   j) NHEBP   k) RLBHE

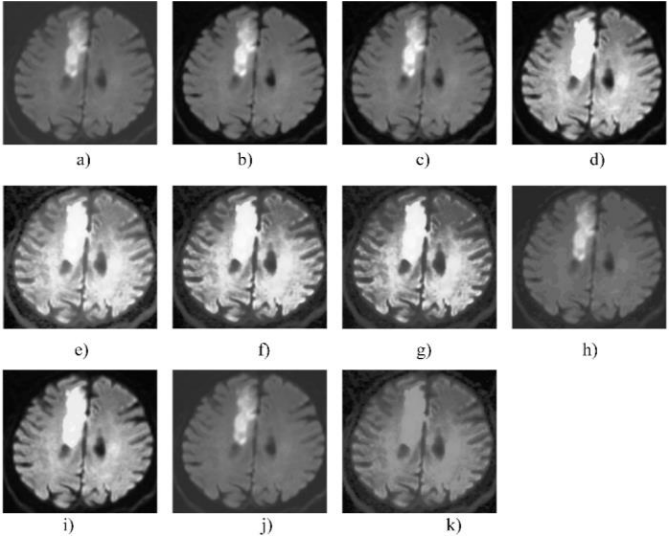

Figure 7: Low Quality Brain MRI a) original Image b) ITSBPL c) MSBPL d) MVSBPL e) HE f) BBHE  g)  DSIHE  h) MMBEBHE   i) BHEPL  j) NHEBP  k) RLBHE

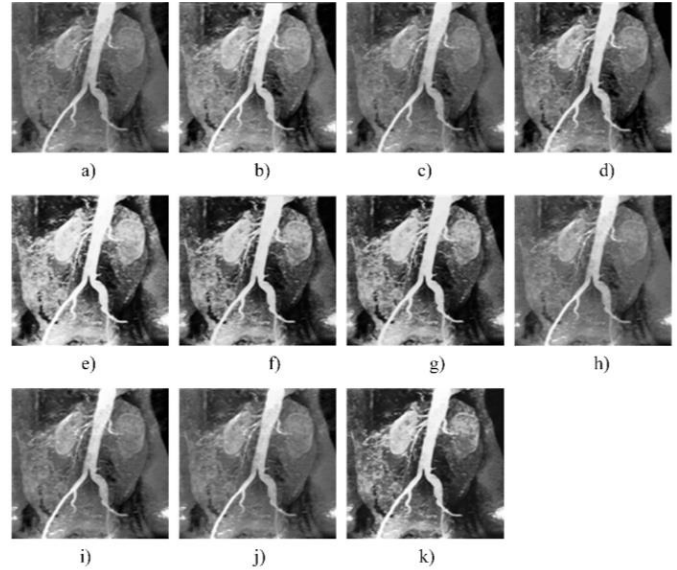

Figure 8:  Kidney image enhancement    a) original Image b) ITSBPL c) MSBPL d) MVSBPL e) HE f) BBHE   g)  DSIHE  h) MMBEBHE   i) BHEPL   j) NHEBP   k) RLBHE

denotes the floor function. It's suggested in application to this parameter that the difference of value of output and input image should be minimum.

$$EME(I) = \underset{k}{argmin}|EME(Y_k) - EME(I_k)| \qquad (32)$$

Hence minimization of enhancement error depends upon the different of input and processed output image [18].

### 4.2.7 Structural Similarity Index (SSIM)
It's the parameter that is being used in measuring image quality by taking the input as original and output as reference image. Luminance, contrast, structure, are the three term that are used to compute the SSIM term. The multiplication of these terms is collective SSIM.

Where $\mu_I, \mu_Y$, $\sigma_I^2, \sigma_Y^2$ and $\sigma_{IY}$ represents the local mean values, standard deviations, and cross-covariance for images $I$ and $Y$.

$$SSIM(I,Y) = \frac{(2\mu_I\mu_Y + C_1)(2\sigma_{IY} + C_2)}{(\mu_I^2 + \mu_Y^2 + C_1)((\sigma_I^2 + \sigma_Y^2 + C_2))} \qquad (33)$$

$C_1 = (K_1L)^2$, $C_2 = (K_2L)^2$, where $K_1, K_2 \ll 1$ and $L$ is the sorted collection of values between 0 to 255 for an image [19]. Better enhancement with values less than one.

For the objective quality analysis, there are total seven tables have presented below that are used to judge the visual quality of the image based on the parameters to measure. Images are in following order brain MRI Slice with tumor, Head MRI with tumor, low quality brain MRI, and Kidney etc. Following are the tables showing the statistical data measure of the visual performance of image enhancement. Seven statistical parameter AMBE, SD, Entropy, PSNR, UIQI, EME, SSIM, etc. have been used to for comparative performance of the algorithms.

Evaluating the results of AMBE states that lower the mean brightness, better is the enhancement, as showed in the Table-1. Value of AMBE found less than other methods because of the robustness of the algorithm in preserving the mean brightness which is evident from the Fig.5 (b)(c)(d). Whereas other algorithms preserve brightness but to some extent, e.g MMBEBHE showing better performance in this case of medical images. Higher the standard deviation better is the enhancement of the target image, which is evident from the Table-2. Average of SD is found for these three criterion are 48-54 which is nominal and counts in good enhancement. For entropy[15], higher the entropy better is the enhancement results which are shown in Table-3. Hence our proposed enhancement criterion proved better as compared to existing techniques. PSNR [16] considered between 20-30db which is attainable using threshold searching bi-histogram method.

Table-4 reveals PSNR values in a specified range. SSI and UIQI are the assessment parameters which satisfy if the enhancement results are closer to unity, Table-5 and Table-7 which show that almost each enhancement method is tending towards unity. Whereas the EME value is lowered for each medical image in Table-6. Hence the above discussion and visual enhancement results prove that ITSBPL, MSBPL, MVSBPL are superior although their comparative study reveals that MVSBPL is much better in enhancement than ITSBPL and MSBPL.

Table 1: Absolute Mean Brightness Error (AMBE)

| Methods | BrainMR | LowQuality | HeadMR | Kidney |
|---|---|---|---|---|
| **ITSBPL** | **0.81** | **0.28** | **0.572** | **0.89** |
| **MSBPL** | **0.34** | **0.38** | **0.496** | **0.39** |
| **MVSBPL** | **0.65** | **0.78** | **0.14** | **0.34** |
| HE | 8.7 | 2.009 | 3.0908 | 1.0081 |
| BBHE | 3.66 | 0.99 | 2.09 | 1.4 |
| DSIHE | 2.5 | 4.74 | 2.97 | 0.391 |
| MMBEBHE | 1.05 | 0.955 | 3.9 | 0.585 |
| BHEPL | 1.94 | 0.4814 | 2.59 | 0.951 |
| NHEBP | 5.44 | 0.767 | 1.026 | 2.85 |
| RLBHE | 1.78 | 3.68 | 2.23 | 1.168 |

Table 2 : Standard Deviation

| Methods | BrainMR | LowQuality | HeadMR | Kidney |
|---|---|---|---|---|
| **ITSBPL** | **43.92** | **50.04** | **60.96** | **49.59** |
| **MSBPL** | **39.39** | **48.48** | **59.66** | **47.23** |
| **MVSBPL** | **43.15** | **51.8** | **67.2** | **55.95** |
| HE | 25.6 | 28.72 | 51.07 | 59.52 |
| BBHE | 41.07 | 55.6 | 65.97 | 76.14 |
| DSIHE | 42.14 | 53.02 | 64.09 | 75.47 |
| MMBEBHE | 40.74 | 45.65 | 37.166 | 42.42 |
| BHEPL | 49.28 | 52.32 | 75.42 | 52.15 |
| NHEBP | 49.37 | 47.19 | 38.68 | 46.7 |
| RLBHE | 4.82 | 29.31 | 44.93 | 64.92 |

Table 3: Entropy

| Methods | BrainMR | LowQuality | HeadMR | Kidney |
|---|---|---|---|---|
| **ITSBPL** | **5.89** | **6.46** | **5.99** | **7.12** |
| **MSBPL** | **5.64** | **6.64** | **5.78** | **7.17** |
| **MVSBPL** | **5.77** | **6.44** | **5.79** | **6.89** |
| HE | 5.4 | 6.1 | 5.82 | 6.89 |
| BBHE | 1.74 | 1.9 | 1.55 | 1.54 |
| DSIHE | 1.79 | 1.76 | 1.52 | 1.53 |
| MMBEBHE | 3.16 | 4.54 | 3.64 | 5.068 |
| BHEPL | 1.87 | 1.75 | 1.64 | 1.08 |
| NHEBP | 6.17 | 6.77 | 6.09 | 7.28 |
| RLBHE | 5.42 | 5.69 | 5.82 | 6.98 |

Table 4: Peak Signal to Noise Ratio (PSNR)

| Methods | BrainMR | LowQuality | HeadMR | Kidney |
|---|---|---|---|---|
| **ITSBPL** | **33.95** | **34.38** | **29.48** | **30.23** |
| **MSBPL** | **31.83** | **39.43** | **29.8** | **34.34** |
| **MVSBPL** | **33.21** | **31.24** | **26.7** | **24.62** |
| HE | 22.43 | 19.46 | 20.13 | 20.46 |
| BBHE | 25.3 | 21.22 | 15.57 | 15.1 |
| DSIHE | 26.78 | 19.62 | 15.8 | 15.2 |
| MMBEBHE | 31.79 | 39.02 | 34.34 | 40.7 |
| BHEPL | 25.79 | 25.8 | 14.54 | 19.77 |
| NHEBP | 28.7 | 46.63 | 40.83 | 33.88 |
| RLBHE | 22.11 | 18.84 | 21.3 | 19.64 |

Table 5: Universal image Quality Index (UIQI)

| Methods | BrainMR | LowQuality | HeadMR | Kidney |
|---|---|---|---|---|
| **ITSBPL** | **0.865** | **0.966** | **0.73** | **0.933** |
| **MSBPL** | **0.465** | **0.995** | **0.752** | **0.95** |
| **MVSBPL** | **0.799** | **0.938** | **0.651** | **0.901** |
| HE | 0.585 | 0.829 | 0.853 | 0.84 |
| BBHE | 0.5944 | 0.579 | 0.674 | 0.711 |
| DSIHE | 0.612 | 0.51 | 0.686 | 0.72 |
| MMBEBHE | 0.634 | 0.838 | 0.99 | 0.99 |
| BHEPL | 0.58 | 0.7188 | 0.586 | 0.866 |
| NHEBP | 0.992 | 0.998 | 0.998 | 0.99 |
| RLBHE | 0.606 | 0.778 | 0.906 | 0.905 |

Table 6: Enhancement Error (EME)

| Methods | BrainMR | LowQuality | HeadMR | Kidney |
|---|---|---|---|---|
| **ITSBPL** | **0.67** | **1.49** | **0.73** | **1.07** |
| **MSBPL** | **0.21** | **1.516** | **0.15** | **0.698** |
| **MVSBPL** | **0.338** | **0.515** | **1.68** | **0.671** |
| HE | 2.47 | 0.594 | 1.642 | 2.16 |
| BBHE | 5.63 | 4.75 | 0.544 | 1.22 |
| DSIHE | 5.62 | 4.94 | 0.54 | 0.72 |
| MMBEBHE | 4.16 | 1.08 | 1.048 | 0.17 |
| BHEPL | 5.73 | 4.75 | 0.569 | 1.87 |
| NHEBP | 0.348 | 0.029 | 0.244 | 0.116 |
| RLBHE | 1.983 | 0.4821 | 1.44 | 0.905 |

Table 7: Structure similarity index (SSI)

| Methods | BrainMR | LowQuality | HeadMR | Kidney |
|---|---|---|---|---|
| **ITSBPL** | **0.949** | **0.989** | **0.788** | **0.9535** |
| **MSBPL** | **0.821** | **0.995** | **0.76** | **0.979** |
| **MVSBPL** | **0.9093** | **0.978** | **0.581** | **0.89** |
| HE | 0.615 | 0.811 | 0.68 | 0.75 |
| BBHE | 0.657 | 0.494 | 0.345 | 0.317 |
| DSIHE | 0.689 | 0.397 | 0.343 | 0.329 |
| MMBEBHE | 0.798 | 0.945 | 0.897 | 0.961 |
| BHEPL | 0.742 | 0.659 | 0.328 | 0.427 |
| NHEBP | 0.9831 | 0.997 | 0.998 | 0.994 |
| RLBHE | 0.606 | 0.764 | 0.76 | 0.794 |

## 5. ACKNOWLEDGEMENT

This work was supported by the National Natural Science Foundation of China (NSFC) under Grant 61271044, 61227802, 61379082, and 61100129.

## 6. CONCLUSION
In this paper, we propose a new bi-histogram equalization method to address the problem of brightness shift and image contrast. The significance of the algorithm is the segmentation of histogram by three different ways. Histogram segmentation divides the histogram into two parts based on searching, mean and Otsu threshold values respectively. Histogram clipping process trims the respective segmented histogram based on calculated platuea limits. Finally, the clipped sub-histograms are equalized independently using histogram equalization. Results reveal that the algorithm perform better in visual and quantitative performance that is evident from the above tables and visually looking at the images. More specifically, MVSBPL is best for brightness preservation and contrast enhancement where ITSBPL and MSBPL ranked second and third respectively.
- It has been tested on medical data that our algorithm preserve 17.1% more brightness than existing techniques.
- Whereas the entropy measure has satisfied richness of information contents by increasing entropy upto 23.9% over existing methods.
- Similarly looking at images visually, could help analysis of texture, feature tracking, segmentation etc.